%% file: metric.tex
\documentclass{article} % For LaTeX2e
\usepackage{a4wide,hyperref,times,graphicx,subfigure,algorithm,algorithmic,appendix}
\usepackage[numbers]{natbib}
\input{math_definition.tex}

\title{
Learning Discriminative Metrics \\via Generative Models and Kernel
Learning}

\author{
Yuan Shi\\
Department of Computer Science\\
University of Southern California\\
Los Angeles, CA 90089\\
\texttt{yuanshi@usc.edu} \\
\and
Yung-Kyun Noh\\
Department of Electrical and Systems Engineering\\
University of Pennsylvania\\
Philadelphia, PA\\
\texttt{nohyung@seas.upenn.edu} \\
\and
Fei Sha\\
Department of Computer Science\\
University of Southern California\\
Los Angeles, CA 90089\\
\texttt{feisha@usc.edu} \\
\and
Daniel D. Lee\\
Department of Electrical and Systems Engineering\\
University of Pennsylvania\\
Philadelphia, PA\\
\texttt{ddlee@seas.upenn.edu} \\
}

\begin{document}

\maketitle

\begin{abstract}
\input{abs.tex}
\end{abstract}

\input{intro.tex}

\input{background.tex}

\input{algo.tex}

\input{exp.tex}

\input{discuss.tex}

\small{
\bibliographystyle{unsrt}
\bibliography{metric,kernel}
}
\newpage
\appendix
\appendixpage

\input{metric_supplement_content.tex}
\end{document}

%% file: math_definition.tex
\usepackage{amsmath,amsfonts}
\usepackage{amsthm,amssymb,amsopn}
\usepackage{bm} % bold symbol

\newcommand{\vct}[1]{\boldsymbol{#1}} % vector
\newcommand{\mat}[1]{\boldsymbol{#1}} % matrix

\newcommand{\field}[1]{\mathbb{#1}}
\newcommand{\R}{\field{R}} % real domain
\newcommand{\T}{^{\textrm T}} % transpose
\newcommand{\TN}{^{-\textrm T}} % transpose

\newcommand{\cst}[1]{\mathsf{#1}}

\newcommand{\twonorm}[1]{\|#1\|_2^2}
\newtheorem{thm}{Theorem}

\newcommand{\eat}[1]{}

%% file: abs.tex
Metrics specifying distances between data points can be learned in a discriminative manner or from generative models.  In this paper, we show how to unify generative and discriminative learning of metrics via a kernel learning framework.  Specifically, we learn local metrics optimized from parametric generative models. These are then used as base kernels to construct a global kernel that minimizes a discriminative training criterion.  We consider both linear and nonlinear combinations of local metric kernels. Our empirical results show that these combinations significantly improve performance on classification tasks.  The proposed learning algorithm is also very efficient, achieving order of magnitude speedup in training time compared to previous discriminative baseline methods.

\iffalse
Metrics measuring the distances between data points can be learnt both discriminatively and generatively.  In this paper, we take a unified approach in combining generative and discriminative learning of metrics. Specifically, metrics optimized under generative modeling assumption serve as the basic building blocks for constructing kernels. These kernels are then combined to minimize discriminative training criterion. We consider both linear and nonlinear combinations.  Our extensive empirical study shows that our generative-discriminative approaches significantly improve performance on classification tasks, when compared to very competitive baselines that are either generatively or discriminatively trained. Additionally, on some datasets, our methods achieves speedup in training time by one to two order-of-magnitude. 
\fi

%view that metrics from generative training criterion can be combined discriminatively. To this end, we view metrics  in the framework of multiple kernel learning. 

%We study a popular setting of metric learning where the metric is
%parameterized by a positive (semi)definite matrix. Our work takes a
%kernel learning view on the metric learning problem, which builds
%connections between existing discriminative and generative metric
%learning techniques. Specifically, we propose two approaches for
%combining generative local metrics: linear combination and
%discriminative learning of nonlinear combination of metric-induced
%kernels.

%% file: intro.tex
\section{Introduction}
\label{sIntro}

Metric learning -- learning how to specify distances between data
points -- has been a topic of much interest  in machine learning
recently.  For example, discriminative techniques for metric
learning aim to improve the performance of a classifier, such as the
$k$-nearest neighbor classifier, on a training set.  As a general
strategy, these techniques try to reduce the distances between data
points belonging to the same class while increasing the distances
between data points from different classes
\cite{Davis07Information,Globerson05Metric,Goldberger05Neighbourhood,Shen09Positive,weinberger09distance,Frome06Image,Bengio09Learning,chopra05learning}.
In this framework, a Mahalanobis metric is parameterized by a
positive (semi)definite matrix, and metric learning is performed
using semi-definite programming (SDP) involving constraints between
pairs or triplets of data points in the training set \cite{boyd}.

\iffalse
Meaningful metrics are useful to nearest neighbor (NN) classifiers.
Examples of such classifiers include the conventional $k$-NN
classifier and kernel classifiers where the kernel functions are
based on distances.

Of particular interests are the discriminative techniques for metric learning
that aim to improve classification accuracies.
As a general strategy, many such techniques try to reduce the distances
between data points which belong to the same class and increase the
distances between data points which are from different
ones~\cite{Davis07Information,Globerson05Metric,Goldberger05Neighbourhood,Shen09Positive,weinberger09distance,Frome06Image,Bengio09Learning,chopra05learning}.
In the popular setting where the metric is parameterized by a
positive (semi)definite matrix, the metric is called Mahalanobis metric. Learning the metric  is often identified as the solution to a semidefinite programming (SDP) with positive semidefinite constraints on the matrix. While there is no closed-form solutions in general, as a special
class of convex optimization, SDPs can be solved efficiently
~\cite{boyd}. As such, learning such metrics is also computationally
appealing.
\fi

% Useful paragraph, just not the right place to put them
%Mahalanobis metric can be used straightforwardly in NN classifiers. Particularly, the squared root of the metric defines a linear mapping which transforms the original data into a new space.  Thus, distances computed using the Mahalanobis metric in the original space are equivalent to the Euclidean distances in the new space.

In the asymptotic limit, the performance of nearest neighbor classifiers approach a theoretical limit, bounded by twice the Bayes optimum error rate, which is independent of the underlying metric used \cite{Duda00Pattern}.  Only in the finite sampling case does the performance of a nearest neighbor classifier depend upon the choice of a metric, and \cite{noh10local} showed how the bias term can be estimated using simple class-conditional generative models fit to the data. A ``generative local metric'' (GLM) is then optimized to minimize this bias term.

However, the local metric learning algorithm has several shortcomings.  First, a local metric needs to be computed at every point, and it is difficult to calculate the geodesic distance between pairs of distant points.  It is also unclear how to correlate the choice of generative models with discriminative classifier performance.

In this paper, we address these issues by combining the learned local metrics in a global discriminative kernel, thus reducing the computational costs for classifying points.  Our approach can be viewed as using metric learning to define base kernels which are then combined discriminately \cite{scholkopf01,lanckriet04kernel,bach04conic,cortes09nonlinear,argyriou05convex}.
The base kernels are derived from parametric generative models, thus reaping the benefits of both generative and discriminative models
\cite{Jaakkola98Exploiting,Ng01On}.
 We show how both simple linear and nonlinear combinations %using Gaussian radial basis functions of learned metrics
result in a highly discriminative global kernel that outperforms  competing methods significantly  on a number of machine leaning datasets.  Moreover, we show that our approach is also computationally more efficient than those methods, often achieving orders of magnitude speedup in training time.

The paper is organized as follows. In section~\ref{sBackground}, we review previous discriminative and generative metric learning techniques.  We describe our approach of combining local metrics trained from generative models in section~\ref{sAlgo}. We present extensive empirical studies in section ~\ref{sExp}, followed by a discussion of our method  and future direction in section~\ref{sDisc}.

The Appendix for this paper includes details of derivations and
implementation, more comprehensive empirical results, and
applications of our approaches to unsupervised learning problems.

%% file: background.tex
\section{Background}
\label{sBackground}

Here we briefly review techniques for learning metrics.  We start with discriminative metric learning,
using the large margin nearest neighbor (LMNN) classifier as an illustrative example~\cite{weinberger09distance}.
 Next, we examine learning a generative local metric (GLM)~\cite{noh10local}, which exploits information from parametric generative models and does not explicity attempt to minimize classification errors.

\subsection{Discriminative learning metric}

Consider a nearest neighbor classifier which labels a
$\cst{D}$-dimensional data point $\vct{x} \in \R^{\cst{D}}$ by the
label(s) of its nearest neighbor(s) $\vct{x}^{NN} \subset
\mathcal{D}$ in a supervised training set $\mathcal{D}$.  In order
to identify the ``nearest'' neighbors, distances from $\vct{x}$ to
data points in $\mathcal{D}$ need to be determined.

The conventional Euclidean distance,
$\twonorm{\vct{x} - \vct{x}'}$, is a special case of the more
general Mahalanobis distance
\begin{equation}
d_{\mat{M}}^2(\vct{x},\vct{x}') = (\vct{x}-\vct{x}')\T\mat{M}(\vct{x}-\vct{x}')\ ,
\label{eMaha}
\end{equation}
when the Mahalanobis metric $\mat{M}\in\R^{\cst{D}\times \cst{D}}$ is equal to the $\cst{D}$-dimensional identity matrix.  In this paper, we follow the popular terminology in the metric learning literature, calling the squared distance as ``distance.''

For a general positive semidefinite matrix
$\mat{M}$,  we can factor it as $\mat{M} = \mat{L}\T\mat{L}$.  This implies a general Mahalanobis metric can be interpreted as Euclidean distance in a transformed space,
$\vct{x} \rightarrow \mat{L}\vct{x}$:
\begin{equation}
d_{\mat{M}}^2(\vct{x},\vct{x}') = (\mat{L}\vct{x}-\mat{L}\vct{x}')\T(\mat{L}\vct{x}-\mat{L}\vct{x}')=\twonorm{\mat{L}\vct{x}-\mat{L}\vct{x}'}.
\label{eMaha2}
\end{equation}

Arguably, the performance of nearest neighbor classifiers depends critically on
the metric $\mat{M}$.  A good $\mat{M}$ should intuitively ``pull'' data points in the same class closer and ``push'' data points in different classes away. This is the general criteria for most discriminative methods for metric learning
\cite{Davis07Information,Globerson05Metric,Goldberger05Neighbourhood,Shen09Positive,weinberger09distance}.  For example, the large margin nearest neighbor
(LMNN) classifier casts the
learning of $\mat{M}$ as a convex optimization problem. For any point $\vct{x}_i$ in the training set, it differentiates two sets of neighboring data points: ``target'' points $\vct{x}_i^+$ whose labels are the same as $\vct{x}_i$  and ``impostor'' points $\vct{x}_i^-$ whose labels are different from $\vct{x}_i$.
LMNN identifies the optimal $\mat{M}$ as the solution to,
\begin{equation}
\begin{aligned}
& \min_{\mat{M} \succeq 0, \, \vct{\xi} \geq 0 }  & &  \sum_i \sum_{j \in \vct{x}_i^+} d_{\mat{M}}^2( \vct{x}_i, \vct{x}_j) + \gamma \sum_{ijl} \xi_{ijl}\\
&  \mathsf{subject\ to}\ \   & &   1 +  d_{\mat{M}}^2( \vct{x}_i,
\vct{x}_j) -  d_{\mat{M}}^2( \vct{x}_i, \vct{x}_l) \le \xi_{ijl},
\forall\ j \in \vct{x}_i^+,\ l \in \vct{x}_i^-
\end{aligned}
\label{eLMNN}
\end{equation}
where the objective function balances two forces: pulling the
targets towards $\vct{x}_i$ and pushing the impostors away so that
the distance to an impostor should be greater than the distance to
the target by a minimum margin of one using the slack variables
$\xi_{ijl}$.

Note that this formulation of LMNN makes no assumptions on how the
(training) data is distributed.  Additionally, the optimization
criterion is directly related to how the learned metric will be used
for classification.  We see that this approach contrasts sharply
with the generative model approach which we describe next.

\subsection{Generative learning metric}

Here we consider a binary classification problem with labels
$y=1,2$, and assume the $\cst{N}$ training data points are drawn
from two class conditional  distributions $p_1(\vct{x}) = p(\vct{x}|
y=1)$ and $p_2(\vct{x}) = p(\vct{x}|y=2)$. In the asymptotic limit,
$\cst{N}\rightarrow \infty$, the error rate of the nearest neighbor
classifier is given in terms of the class conditional
distributions\footnote[1]{For simplicity, we consider equal prior
distributions here. Unequal class priors contribute a more
complicated scalar term in Eqs.
(\ref{eAsymp}-\ref{eq:MetricEffect}), but the resulting derivation
for the optimization of the local metric is unchanged.}:
\begin{equation}
{\varepsilon}_{\infty} =  \int\frac{p_1(\vct{x})p_2(\vct{x})}{p_1(\vct{x})+p_2(\vct{x})}d\vct{x} .
\label{eAsymp}
\end{equation}
The asymptotic error $\varepsilon_\infty$ can be easily shown to be
invariant to a linear transformation of variables: $\vct{z} =
\mat{L}\vct{x}$. This implies that learning a different metric
$\mat{M}$ in Eq.~(\ref{eMaha2}) should have no effect on the error
rate in the asymptotic limit.

The solution to this apparent paradox is described in \cite{noh10local}, which showed that when the number of training points is finite, the error rate of the nearest neighbor classifier deviates from the asymptotic error rate
$\varepsilon_\infty$ by a finite bias term,
\begin{equation}
\varepsilon_{\cst{N}}  \simeq \varepsilon_\infty
 + c_{\cst{N}}\int\frac{p_1p_2(p_2-p_1)}{(p_1 +
p_2)^2}\left[\frac{\nabla^2p_1}{p_1} - \frac{\nabla^2p_2}{p_2} \right]d\vct{x} \label{eNN}
\end{equation}
where the constant factor $c_{\cst{N}}$ tends to zero as
$\cst{N}$ approaches infinity, and the scalar
Laplacian $\nabla^2p(\vct{x})$ is the trace of the Hessian
$\nabla\nabla p (\vct{x})$.

This bias term \emph{does} depend upon the choice of the underlying metric, and
under a linear transformation
$\vct{z}=\mat{L}\vct{x}$, the bias term is given by the integral
of
\begin{equation}
\label{eq:MetricEffect}
\frac{p_1p_2(p_2-p_1)}{(p_1 + p_2)^2}\mathsf{Trace}\left[\mat{M}^{-1}\mat{\Phi}\right], \mbox{ with }
\mat{\Phi}  = \frac{\nabla\nabla
p_1(\vct{x})}{p_1(\vct{x})} - \frac{\nabla\nabla
p_2(\vct{x})}{p_2(\vct{x})}.
\end{equation}
The generative local metric (GLM) algorithm optimizes a local metric $\mat{M}$ to minimize the local bias term in eq.~(\ref{eq:MetricEffect}), so that the finite sample error rate
$\varepsilon_\cst{N}$ approaches the asymptotic error rate $\varepsilon_\infty$ to
the first-order approximation.  The resulting optimization with semidefinite constraints:
\begin{equation}
\boxed{
\min_{\mat{M}_i}\ \  \left( \mathsf{Trace}\left[\mat{M}_i^{-1}\mat{\Phi}_i\right]\right)^2,\ \ \mathsf{subject\ to}\ \    |\mat{M}_i|=1,\ \mat{M}_i \succeq 0}
\label{eLocalSDP}
\end{equation}
is easily solved at each data point $\vct{x}_i$ using a spectral
decomposition. The optimum $\mat{M}_i^*$ is a positive semidefinite
matrix whose eigenvectors $\mat{U}_i$ are the same as
$\mat{\Phi}_i$'s. Then if $\mat{\Lambda}^+$ is the diagonal matrix
composed of $\mat{\Phi}_i$'s $\cst{d^+}$ positive eigenvalues, and
$\mat{\Lambda}^-$ is the corresponding diagonal matrix with
$\cst{d^-}$ negative eigenvalues, the solution can be written as:
\begin{equation}\label{eLocal}
\mat{M}_i^* \propto  \mat{U}_i\left[ \begin{aligned}
\cst{d^+} \mat{\Lambda}^+ &\\
& -\cst{d^-} \mat{\Lambda}^-\end{aligned}\right]\mat{U}_i\T
\end{equation}
where the proportionality constant is determined by scaling
the
determinant of $\mat{M}_i^*$ to unity.  Note that this learning
algorithm does not attempt to reduce the nearest neighbor classification error rate explicitly.

%% file: algo.tex
\section{Discriminative learning with multiple generative metrics }
\label{sAlgo} Prior empirical studies have shown that generative
learning metric  (GLM) of eq.~(\ref{eLocal}) performs competitively,
even when compared to discriminative methods such as the large
margin nearest neighbor classifiers
(LMNN)~\cite{weinberger09distance}. However, GLM has has several
shortcomings. For every (new) data point $\vct{x}$ to be classified,
the optimization of eq.~(\ref{eLocal}) needs to be solved. The
resulting metric $\mat{M}_{\vct{x}}$ depends on $\vct{x}$ and
distances to the training data points need to be computed in this
specific metric. While specialized data structures can be exploited
to speed up the process of identifying nearest neighbors, these
structures usually require a fixed metric and cannot be easily
adapted to a new metric~\cite{weinberger09distance}. This can
significantly increase the computational cost at testing time.

Secondly, the performance of GLM depends on the specific form of the
class conditional distributions $p_1(\vct{x})$ and $p_2(\vct{x})$
used for generative modeling.  Initial studies have suggested that
even with simplistic models such as Gaussian distributions, GLM
attains robust and competitive performance.  Nevertheless,
rigorously quantifying the relationship between the assumed
generative models and the classification performance is lacking.  In
particular,  it is unclear how the choice of the generative models
should be adapted in order to further improve the classification
performance of the nearest neighbor classifier with the learned
metric.

We address these issues by viewing the problem of learning metrics as learning kernels.  We then investigate how to improve classification performance by using these kernels. To this end, we consider two schemes for learning kernels discriminatively: linear and nonlinear combination of metrics.

\subsection{Linear combination of local metrics}
\label{sConvex}

A metric $\mat{M}_i$ learned at the training point $\vct{x}_i$ can be seen as a linear positive semidefinite kernel, defining the inner product between two points $\vct{x}_m$ and $\vct{x}_n$,
\begin{equation}
K_i( \vct{x}_m, \vct{x}_n) = \vct{x}_m\T \vct{M}_i \vct{x}_n
\end{equation}
Note that while $\vct{M}_i$ is learned ``locally'' in the neighborhood of $\vct{x}_i$, we treat it as an biased estimate of a \emph{global} kernel function over the space of
all training examples.
We arrive at an unbiased estimator of the global metric  --- intuition to be made clear below --- by linearly combining all the local kernels learned from the $\cst{N}$ training samples,
\begin{equation}
K(\vct{x}_m, \vct{x}_n) = \sum_i \alpha_i K_i( \vct{x}_m, \vct{x}_n) =  \vct{x}_m\T \left(\sum_i \alpha_i \mat{M}_i  \right) \vct{x}_n = \vct{x}_m\T  \mat{M}\vct{x}_n
\label{eConvex}
\end{equation}
where the combination coefficients $\{\alpha_i\}_{i=1}^{\cst{N}}$
are constrained to be nonnegative and sum to one, guaranteeing the
resulting kernel to be positive semidefinite. The global metric
$\mat{M}$ is then simply the convex combination of all local
metrics.  We now consider the simplest convex combination, uniform
averaging:
\begin{equation}
\boxed{
\mat{M}^{\textrm{UNI}} = \frac{1}{\cst{N}} \sum_i \mat{M}_i}\ .
\label{eUniform}
\end{equation}
As our empirical studies show, this
surprisingly simple strategy works well
in practice.

% We consider two
%sets of combining coefficients.
%
%\begin{itemize}
%\item[\textbf{Uniform}] We simply set $\alpha_i = 1/N$ where $N$ is the number of training examples.
%As the empirical study will show later, this surprisingly simple strategy works well in practice.
%
%\item[\textbf{Weighted}] We weight each metric by the density of $\vct{x}_i$, estimated from a density estimator.
% Note that this density estimator does not have to be the same as  with the generative model assumption of the data used for generative metric learning.  For instance, in our empirical study, we assume that each class conditional distribution $p(\vct{x}| c)$ is a Gaussian distribution. We then use a kernel density estimator to compute $\alpha_i$.
%\end{itemize}

As noted earlier, a positive semidefinite metric may be viewed as applying a linear transformation to the original data $\vct{x}$.  This implies that $\mat{M}^{\textrm{UNI}}$ transforms $\vct{x}$ to a new space where on average, local metrics computed in that space are proportionally to the identity matrix.  Thus, on average, the Euclidean distance based nearest neighbor classification will perform well in that space. More formally,

\begin{thm}
\label{thUniform}
Assume the class conditional distribution $p_c(\vct{x}) = p(\vct{x}|y= c)$ is Gaussian for every class. Let $\mat{M}_i$ be the local metric computed with eq.~(\ref{eLocalSDP}), minimizing the bias term in the space of $\vct{x}$. Then, the uniform convex combination metric $\mat{M}^{\textrm{UNI}}$ of eq.~(\ref{eUniform}) induces a linear transformation $\vct{z} = \mat{L}\vct{x}$ where $\mat{L}\T\mat{L} = \mat{M}$. Furthermore, let $\mat{Q}_i$ denote the local metric computed in the space of $\vct{z}$ under the new class conditional distribution $p_c(\vct{z})$. We have,
\begin{equation}
\sum_{i=1}^{\cst{N}} \mat{Q}_i \propto \mat{I}
\end{equation}
where $\mat{I}$ is the identity matrix.
\end{thm}
The proof exploits the fact that $p_c(\vct{z})$ is also Gaussian and
$\mat{Q}_i$ can be expressed as a closed-form expression of
$\mat{M}_i$.  Details are presented in the Appendix
(section~\ref{sProof}).

\subsection{Nonlinear combination of local metrics}
\label{sNonlinear}

In order to combine local metrics in a nonlinear fashion,  we use Gaussian radial basis (RBF) kernel functions to replace the standard identity covariance matrix,
\begin{equation}
K_{il}(\vct{x}_m, \vct{x}_n) = \exp \left\{-(\vct{x}_m-\vct{x}_n)\T \mat{M}_i
(\vct{x}_m-\vct{x}_n)/\sigma_l^2 \right\}
\label{eNonlinear}
\end{equation}
where $\sigma_l$ is a bandwidth parameter, chosen from a range of possible values between $\sigma_{\min}$ and $\sigma_{\max}$.

Our goal is to learn a convex combination of these RBF base kernels. We follow the standard multiple kernel learning (MKL) framework where base kernels are combined as~\cite{lanckriet04kernel},
\begin{equation}
K(\vct{x}_m, \vct{x}_n) = \sum_{i,l} \alpha_{il} K_{il}(\vct{x}_m, \vct{x}_n)\ \ \mathsf{subject\ to}\ \alpha_{i,l} \geq 0,\  \sum_{i,l} \alpha_{i,l} = 1
\label{eMKL}
\end{equation}
Note that the combined kernel $K(\cdot, \cdot)$ is a highly
nonlinear, albeit convex function of local metrics.  It is
well-known that positive semidefinite kernels, including ours in
eq.~(\ref{eMKL}), can be represented as distances in the
corresponding Reproducing  Kerner Hilbert Space
(RHKS)~\cite{scholkopf00kernel}. However, as opposed to the global
metric $\mat{M}^{\textrm{UNI}}$, we cannot represent this distance
(and its associated metric) as a closed-form function of $\vct{x}$
and $\{\mat{M}_i\}_{i=1}^{\cst{N}}$.

In typical applications of MKL, one often chooses Gaussian RBF kernels with identity covariance matrices $\exp(-\twonorm{\vct{x}_m-\vct{x}_n}/\sigma_l^2)$.  This is due to the difficulty in properly choosing non-identity covariance matrices for the base kernels, especially in high-dimensional problems.  Our formulation in eq.~(\ref{eMKL}) overcomes the challenge by using non-Euclidean metrics computed from generative modeling.

We refine the combination by optimizing $\{\alpha_{il}\}$ discriminatively.  Specifically, the coefficients $\{\alpha_{il}\}$ are adjusted so that the kernel $K(\cdot, \cdot)$ achieves the lowest empirical risk when used in kernel based classifiers such as support vector machines~\cite{lanckriet04kernel}.  In this aspect, our formulation reaps the benefits of both generative modeling and discriminative training.

\subsection{Convex combination: revisited}
\label{sOther} One may wonder why the framework of multiple kernel
learning, used for nonlinear combination of metrics in
section~\ref{sNonlinear}, is not used to discriminatively optimize
the convex combination coefficients of eq.~(\ref{eConvex}).  Our
preliminary results indicate that $\mat{M}^{\textrm{UNI}}$ in
general performs well. This is \emph{consistent} with previous
extensive work on combining kernels linearly --- discriminative
learning of such combinations does not reliably outperform simpler
strategies of combinations including the uniform
combination~\cite{cortes09l2,lanckriet04kernel}.  We present more
experimental details, including other forms of convex combinations,
in the Appendix (section~\ref{sKDE}).  We have found that
$\mat{M}^{\textrm{UNI}}$ is both computationally appealing and
empirically very effective.

\subsection{Computational complexity and optimization}

The computational complexity of our algorithms is dominated by the calculation of local metrics in eq.~(\ref{eLocal}).  The main calculation involved is diagonalizing the matrix $\mat{\Phi}$.  For $\cst{D}$-dimensional space $\vct{x}$, the computational cost is $O(\cst{D}^3)$. Since the local metrics are computed at every training sample, the total computational cost is $O(\cst{N}\cst{D}^3)$. Computing $\mat{M}^{\textrm{UNI}}$ itself adds little overhead.

In contrast, discriminative techniques such as LMNN for learning a single global metric require iterative numerical optimization. For LMNN, the optimization needs to examine roughly $O(\cst{N}^3)$ number of constraints. For very large $\cst{N}$ and small to moderate $\cst{D} \le \sqrt{\cst{N}}$, our approaches will greatly outperform LMNN in speed, as demonstrated later in section~\ref{sExp}.

%We defer the discussion on the  computational complexity of our nonlinear combination of metrics to the Appendix (section~\ref{sExpDetail}).

%depends on $\cst{N}$, the number of training samples and local metrics, as well as the search range of $\sigma_l$. In practice, to reduce the number of the base kernels we need to consider, we cluster them into few tens to hundreds. This heuristic works well empirically. To learn the coefficients $\alpha_{il}$, we have adopted the SimpleMKL algorithm and used the software package provided by the authors of that algorithm.

%% file: exp.tex
\section{Experimental results}
\label{sExp}

We compare our methods of discriminative kernel learning from
generative local metrics (GLMs) to other competitive methods of
metric learning.  Here we report the results of applying simple
linear (section~\ref{sConvex}) and nonlinear combinations
(section~\ref{sNonlinear}) to classification. More comprehensive
details are included in the Appendix (section~\ref{sKDE} --
\ref{sUn}).

\begin{table}[t]
\caption{Error rates of misclassification (in \%) on the 8 small-scale datasets} \label{tSmall}
\begin{center}
\begin{tabular}{|c|c|c|c|c|c|c|c|c|c|} \hline
\multicolumn{1}{|c|}{\textbf{Method}}  &
\multicolumn{8}{c|}{\textbf{Dataset}} &
\multicolumn{1}{c|}{\textbf{Avg.}}
\\ \cline{2-9} & 3-Norm. & Wine &
Iris & Heart & Vehi. & Ionos. & Image & German & \textbf{Rank} \\
\hline Euclidean & 8.17 & 4.41 & 5.11 & 22.78 & 31.29 & 15.87 & 2.68
& 27.27 & 5.75\\
\hline LMNN & 4.70 & 2.61 & 4.78 & 21.91 & 21.97 & 12.39 &
\textbf{2.14} &
27.20 & 4.25\\
\hline GLM$^{\textrm{INT}}$ & 3.83 & 3.96  & 3.67 &21.60 & 25.32  & \textbf{6.24}  &
2.89  &26.17  & 3.75\\
\hline $\mat{M}^{\textrm{UNI}}$ & 3.44  & \textbf{1.80} & 3.33 &19.51 &17.47  & 9.67 & 2.47  &26.12 & \textbf{1.88}\\
\hline LMNN$_E$ & 3.70  & 2.43 & 4.67 &20.56 &20.37 & 11.97 & 2.67 &26.88 & 3.25\\
\hline $\mat{M}_E^{\textrm{UNI}}$ & \textbf{3.10} & 2.52  & \textbf{3.11} &\textbf{19.26} &\textbf{15.81} & 10.80 & 3.01  &\textbf{25.15 }& 2.13\\
\hline
\end{tabular}
\end{center}
\end{table}

\subsection{Setup}

\textbf{Datasets} We have used 10 datasets: 3-Normal, Iris, Wine, Heart, Vehicle, Ionosphere, Image, German,
MNIST and Letters. The first 8 datasets are small-scale, having
150--2310 data points with dimensionality ranging from 4 to 34. The
number of labelled classes range from $2$ to $4$.  The MNIST and
Letters datasets are substantially larger: MNIST has 70,000 deskewed
images with 10 classes while Letters has 20,000 examples with 26
labelled classes.
% small-scale datasets are  summary statistics of these datasets are listed in
%Table~\ref{tDataset}, where the datasets of MNIST and Letters are considered to be large-scale. From the left to the right in the table, the
%first dataset (
3-Normal is a synthetic set containing a mixture of 3 Gaussians.
Other datasets  are downloaded from  the UCI machine learning
repository~\cite{Frank10uci},
 the IDA benchmark
repository\footnote{http://www.fml.tuebingen.mpg.de/Members/raetsch/benchmark} and NYU
~\cite{lecun98mnist}.

Data in the small-scale datasets is preprocessed so that the feature
vector components range between $-1$ and $1$. For supervised
learning tasks such as classification, each dataset is randomly
split 30 times into training ($60\%$), validation ($20\%$) and
testing ($20\%$) sets.

The MNIST images have a resolution of $784$ pixels and are
preprocessed with PCA, reducing the dimensionality to 40, 60 and 164
respectively, to save training time and prevent overfitting. We
perform 5 random splits, each with 65000 samples for training, 5000
for validation and 10000 for testing.  For the Letters dataset, we
perform 10 random splits, each with 12000 samples for training, 2000
for validation and 6000 for testing. The Letters-scaled set is with
the features scaled to lie within the range $[-1, 1]$. We also
provide experimental results on the unscaled version (denoted as
Letters-original), as the training time of LMNN is sensitive to the
scaling for this dataset.

\textbf{Learning methods} The various learning methods used in our
comparative study are summarized below:
\begin{list}{\labelitemi}{\leftmargin=1em}
  \item Euclidean.  $k$-nearest neighbor ($k$NN) classifier using Euclidean distances.
  \item LMNN.   $k$NN classifier using the metric of Large Margin Nearest Neighbor~\cite{weinberger09distance} (cf. section~\ref{sBackground}).
  \item GLM$^{\textrm{INT}}$.  $k$NN classifier using the generative local metrics (GLM)~\cite{noh10local} (cf. section~\ref{sBackground}).  Gaussian distributions are used as the class-conditional distributions for generative modeling. We follow the procedure in~\cite{noh10local} to interpolate GLM with the Euclidean metric. (Un-interpolated GLMs underperform interpolated ones, a finding which is consistent with what is reported in~\cite{noh10local}).
    \item $\mat{M}^{\textrm{UNI}}$. $k$NN classifier using \textbf{our} approach of  uniformly combining the \emph{un-interpolated} GLMs into a single metric, described by eq.~(\ref{eUniform}).
  \item LMNN$_E$. Energy-based classification using the metric of
    LMNN~\cite{weinberger09distance}\footnote{The energy of a point
      being assigned to a class $c$ is defined as the differences
      between two quantities: the (sum of) distances of this point to
      its nearest neighbor in the class $c$ and the (sum of) distances
      of this point to its nearest neighbor in classes other than
      $c$. A point is assigned to a class label which has the lowest
      energy. Energy-based classification can improve performance significantly over $k$NN classification~\cite{weinberger09distance}.}.
  \item $\mat{M}^{\textrm{UNI}}_E$. Energy-based classification using the metric of  $\mat{M}^{\textrm{UNI}}$.
\end{list}

%We compare G$^2$M-Uni to two state-of-the-art metric learning
%methods - large margin nearest neighbor (LMNN) and generative local
%metric learning (GLML). LMNN is a discriminative method which learns
%a global metric, while GLML is a generative method which learns a
%metric for each data point. We also include a baseline method which
%simply uses the Euclidean distance metric. To evaluate the
%performance of different metrics, we apply $k$-NN classifier on the
%learnt metric. We also apply a variant of $k$-NN classifier -
%energy-based classification method \cite{Weinberger05Distance} on
%the metrics learnt from LMNN and G$^2$M-Uni. We use energy-based
%classification mainly for a fair comparison with LMNN, because it
%has shown that energy-based classification often leads to better
%performance than $k$-NN on LMNN metric.

\textbf{Tuning} The parameters of all methods are optimized on
validation sets, and their overall performance is reported on test sets.
Tunable parameters include the number of (target) nearest neighbors,
the interpolation parameter in the GLM$^{\textrm{INT}}$, and the
margins used in the two energy-based classification. We have used the
LMNN implementation as reported in~\cite{weinberger09distance}.

\subsection{Linear combination of generative local metrics}

\textbf{Performance on classification tasks} Table \ref{tSmall}
displays averaged misclassification error rates (over 30 random
splits)  for the 8 small-scale data sets. Standard errors are
reported in the Appendix (section~\ref{sExpDetail}).  The last
column is  the averaged ranking in performance (across 8 datasets);
the smaller the number  the better the performance is on average.
GLM$^{\textrm{INT}}$ outperforms LMNN and Euclidean on most sets,
though its performance is surpassed by LMNN$_E$. However, the best
performers are $\mat{M}^{\textrm{UNI}}$ and
$\mat{M}_E^{\textrm{UNI}}$, which use the simple strategy of
uniformly combining the generative local metrics.

Table \ref{tLarge} displays averaged error rates (over 5 random
splits for MNIST and 10 for Letters) on large-scale datasets.
Standard errors are reported in the Appendix
(section~\ref{sExpDetail}). For the MNIST dataset,  we report
results on several PCA-preprocessed dimensionality\footnote{ As a
comparison point, MNIST-164 has the same dimensionality as the one
reported in the work of LMNN~\cite{weinberger09distance}, with an
error rate of $1.37\%$ using the energy-based classification,
denoted as LMNN$_E$ in the table. We obtain a similar error rate of
$1.34\%$.}. For Letters, we report results on two cases using scaled
and unscaled features.

On the MNIST dataset, it is clear that both
$\mat{M}_E^{\textrm{UNI}}$ and LMNN$_E$ perform better than other
methods when the dimensionality is low ($\cst{D}\le 60)$. However,
at the larger dimensionality of 164, both LMNN and LMNN$_E$
outperform other methods including our approaches.  One possible
explanation is that, with the increased dimensionality, the
generative modeling used by both GLM$^{\textrm{INT}}$ and our
approaches ($\mat{M}^{\textrm{UNI}}$ and
$\mat{M}_E^{\textrm{UNI}}$), does not fit the data properly.  On the
other hand, discriminative training might be able to overcome the
problem with better regularization.

%This conjecture is supported by our preliminary examination of the LMNN metric at $\cst{D}=164$, which is rank-deficient with the rank being around 40.  It is interesting to notice that at $\cst{D}=40$, the performance of our approach $\mat{M}_E^{\textrm{UNI}}$ is as good as the best performing LMNN$_E$.

On the Letters dataset, our approach of $\mat{M}_E^{\textrm{UNI}}$
clearly outperforms all other methods.

\begin{table}[t]
\caption{Error rates of misclassification (in \%) on the two large-scale datasets} \label{tLarge}
\begin{center}
\begin{tabular}{|c|c|c|c|c|c|} \hline
& MNIST-40 & MNIST-60 & MNIST-164 & Letters-scaled &
Letters-original
\\ \hline
Euclidean       &2.09      & 2.02      & 2.16      &5.05   & 5.25       \\
\hline
LMNN            &1.99       & 1.84     & 1.82      &3.91    & 3.81     \\
\hline
GLM$^{\textrm{INT}}$          &3.75       & 3.55     & 3.48      & 5.55  & 5.51 \\
\hline
$\mat{M}^{\textrm{UNI}}$    &1.93       & 1.90      & 4.30     &3.04  & 2.96\\
\hline
LMNN$_E$     &1.53       & \textbf{1.43}      & \textbf{1.34}    &2.98  & 2.90 \\
\hline
$\mat{M}_E^{\textrm{UNI}}$   &\textbf{1.40}      & 1.44      & 3.13     &\textbf{2.26}  & \textbf{2.28}\\
\hline
\end{tabular}
\end{center}
\end{table}

%\begin{table}[t]
%\caption{Error rates on large-scale datasets} \label{error-large}
%\begin{center}
%\begin{tabular}{|l|c|c|c|c|c|} \hline
%\multicolumn{1}{|c|}{\bf METHOD}  & \multicolumn{5}{c|}{\bf DATASET}
%\\ \hline
%& MNIST (40) & MNIST (60) & MNIST (164) & Letters-scaled & Letters-original \\ \hline
%Euclidean       &2.09 $\pm$ 0.02      & 2.02 $\pm$ 0.02     & 2.16 $\pm$ 0.01     &5.05 $\pm$ 0.07   & 5.25 $\pm$ 0.08       \\
%\hline
%LMNN            &1.99 $\pm$ 0.03      & 1.84 $\pm$ 0.01     & 1.82 $\pm$ 0.03     &3.91 $\pm$ 0.08   & 3.81 $\pm$ 0.13      \\
%\hline
%GLML           &3.75 $\pm$ 0.06      & 3.55 $\pm$ 0.05     & 3.48 $\pm$ 0.09     & 5.55 $\pm$ 0.08  & 5.51 $\pm$ 0.08\\
%\hline
%G$^2$M-Uniform          &1.93 $\pm$ 0.03      & 1.90 $\pm$ 0.03     & 4.30 $\pm$ 0.04     &3.04 $\pm$ 0.08  & 2.96 $\pm$ 0.09\\
%\hline
%LMNN-Energy     &1.53 $\pm$ 0.01      & \textbf{1.43} $\pm$ 0.01     & \textbf{1.34} $\pm$ 0.01     &2.98 $\pm$ 0.06  & 2.90 $\pm$ 0.06\\
%\hline
%G$^2$M-Uniform-Energy   &\textbf{1.40} $\pm$ 0.01      & 1.44 $\pm$ 0.01     & 3.13 $\pm$ 0.03     &\textbf{2.26} $\pm$ 0.04   & \textbf{2.28}$\pm$ 0.06\\
%\hline
%\end{tabular}
%\end{center}
%\end{table}

\textbf{Computational efficiency} Details are presented in the
Appendix (section~\ref{sTime}). In summary, we observe that our
methods are computationally efficient, achieving orders of magnitude
speedup in training time.  For example, on MNIST-40, LMNN takes
about 40 minutes to learn the final metric while our
$\mat{M}^{\textrm{UNI}}$ algorithm takes about 4 minutes.

%We mention briefly the difference in computation times used by LMNN and our methods in learning a metric.

\subsection{Nonlinear combination of generative local metrics}

We also report the results of a nonlinear combination of generative local
metrics, using the framework of discriminative kernel learning described in section~\ref{sNonlinear}.
%We learn the nonlinear combination of metrics in the framework of convex combination of nonlinear kernels~\cite{lanckriet04kernel,rakotomamonjy08simple}.
The baseline system learns a kernel in the following form
$K(\vct{x}, \vct{x}') = \sum_l \alpha_l \exp\{-\twonorm{\vct{x}-
\vct{x}'}/\sigma_l^2\} $ where $\sigma_l$ is the bandwidth of the
kernel. Our method of discriminative kernel learning replaces the
Euclidean distance in the conventional Gaussian RBF kernel with the
generative local metrics (GLMs),  as in eq.~(\ref{eNonlinear}).
There are as many local metrics as the number of training examples.
Thus, for our implementation, we use ``regional'' metrics in
eq.~(\ref{eNonlinear}).  Specifically, we partition the training
data into $\cst{P}$ parts. We average local metrics of data points
in each part and obtain $\cst{P}$ ``regional'' metrics
$\{\mat{M}_p\}_{p=1}^{\cst{P}}$. In the specific case where
$\cst{P}=1$, we will get $\mat{M}^{\textrm{UNI}}$, the uniform
linearly combined metric.

%We then use those `regional'' metrics to compose the desired kernel with the formulation of . We use the same scaling sc
%heme for constructing the baseline eq.~(\ref{eBaseline}) and adjust $\sigma_0^2$ accordingly (with respect to each different regional metric).

For both the baseline method and our approach of kernel learning with
eq.~(\ref{eNonlinear}), the combination coefficients are optimized
with the SimpleMKL
algorithm~\cite{rakotomamonjy08simple}. Table~\ref{tNonlinear}
displays averaged misclassification error rates on 7 (out of 8)
small-scale datasets.  Experiments on other datasets are ongoing.

%We adopt  to optimize the combination coefficients and use the codes provided by its uthors.

We experiment with different $\cst{P}=1, 5,$ and $10$ and aggregate
the results by reporting the best performing $\cst{P}$ in
Table~\ref{tNonlinear}. Further details of our method's performance
with different $\cst{P}$ are provided in the Appendix,
section~\ref{sExpDetail}.  On 5 out of the 7 datasets, nonlinear
combination of metrics clearly outperforms the baseline, with
significant improvement on the datasets of 3-Normal, Iris and
Vehicle; however, our method performs poorly on the Ionosphere
dataset. Note that in Table~\ref{tSmall}, the local metrics used
alone attain a better error rate ($6.24\%$) than the best nonlinear
kernel method ($6.90\%$).  Thus, more analysis is still needed to
understand effective methods for nonlinear combinations of local
metrics.

%We compare 3 types of methods:
%\begin{itemize}
%  \item \textbf{Identity}. The metric $M$ is the identity matrix. It learns the combination of 15 kernels (due to 15 different bandwidth parameters).
%  \item \textbf{Global}. The metric $M$ is from G$^2$M-Uniform. It learns the combination of 15 kernels.
%  \item \textbf{Local($k$)}. This method first clusters the data into $k$ clusters
%under the metric from G$^2$M-Uniform, then for each cluster $c \in
%\{1,...,k\}$, it compute the average of local metrics in that
%cluster (denoted by $G_c$). Then, it learns the combination of
%kernels defined on $\{G_c\}$ (15$k$ kernels in total).
%\end{itemize}

%The regularization parameter $C$ in SimpleMKL is tuned in $[2^{-3},
%2^{-2}, ..., 2^10, 2^{11}]$ on validation set. For Local($k$), we
%set $k$ equal to 1, 5 and 10, respectively, and report the best
%results. Table \ref{error-nonlinear} compares the error rates
%(averaged over 30 random splits) of Identity and Local($k$). On four
%datasets, Local($k$) leads to lower classification error rates than
%Identity, which suggests that the kernels defined on local metrics
%are good candidates for base kernels in multiple kernel learning. On
%Ionosphere and Wine datasets, however, Local($k$) underperforms
%Identity, due to the overfitting of local metrics on the training
%data. One possible way to solve this issue is to learn local metrics
%on a different training set.

\begin{table}[t]
\caption{Error rates of misclassification (in \%) with nonlinear
combination of metrics} \label{tNonlinear}
\begin{center}
\begin{tabular}{|c|c|c|c|c|c|c|c|}\hline
& 3-Normal & Wine & Iris & Heart & Vehi. & Ionos. & German \\ \hline
Baseline & 6.08 & \textbf{2.16} & 5.56 & 18.09 & 26.53 & \textbf{6.90} & 25.22      \\
\hline
Our approach & \textbf{2.31} & 2.43 & \textbf{2.78} & \textbf{17.16} & \textbf{15.44} & 9.01 & \textbf{24.25}   \\
\hline
\end{tabular}
\end{center}
\end{table}

\eat{
\begin{table}[t]
\caption{Error rates of combining nonlinear kernels}
\label{error-nonlinear}
\begin{center}
\begin{tabular}{|c|c|c|c|c|c|c|}\hline
& Ionos. & Heart & 2-Norm. & German & Wine & Iris \\ \hline
Euclidean       &\textbf{6.90}    & 18.09    & 3.17   &25.22   & \textbf{2.16 } & 5.56   \\
\hline
Local ($k$)   &9.01      & 17.16    & 1.53    &24.25 & 2.43  & \textbf{2.78}  \\
\hline
\end{tabular}
\end{center}
\end{table}
}

\subsection{Application to unsupervised learning problems}
\label{sUnsupervised}

Many unsupervised learning problems, such as clustering and
dimensionality reduction, also rely upon a proper metric to
calculate distances. We have also investigated how to apply
algorithms to learn metrics to such unsupervised problems.  One
crucial step is to extract discriminative information from unlabeled
data for the algorithms to compute better metrics.  To address this
issue, we have developed an EM-like procedure to iterate between
inferring labels and computing local and global metrics. Details are
presented in the Appendix (section~\ref{sUn}).

We applied this procedure to a number of unsupervised learning
problems.  We achieve significantly better performance than standard
approaches for clustering.  Additionally, we can exploit the learned
metric for dimensionality reduction, for instance --- learning the
nonlinear manifold structure of data. As an illustrate example, we
show the benefits of using the global metric $\mat{M}^{\textrm{UNI}}$
with the algorithm of IsoMap~\cite{isomap}.
%In unsupervised clustering, we start with K-means clustering with the Euclidean distance metric. We then treat the cluster labels \emph{as if} they are ground-truth class labels. We apply the generative metric learning algorithm to such ``labeled'' data, compute local metrics and then global metric.  We then apply K-means clustering again using the learnt global metric to compute distances.  We iterate the process for a few times or until the cluster labels no longer change. Our empirical study shows that this simple strategy works very well. The algorithm returns with clustering of higher quality, measured in standard measures, than K-means. Note that being able to have a global metric is essential. Without it, it is difficult to compare distances computed from different (local) metrics.

 %As a representative method for manifold learning, Isomap relies on the distance information of the neighborhood, which depends on the distance metric. To illustrate this,

\begin{figure}
  \centering
  \subfigure[ISOMAP + Euclidean]{
    \label{fig:IsomapEuc} %% label for first subfigure
    \includegraphics[width=2.2in]{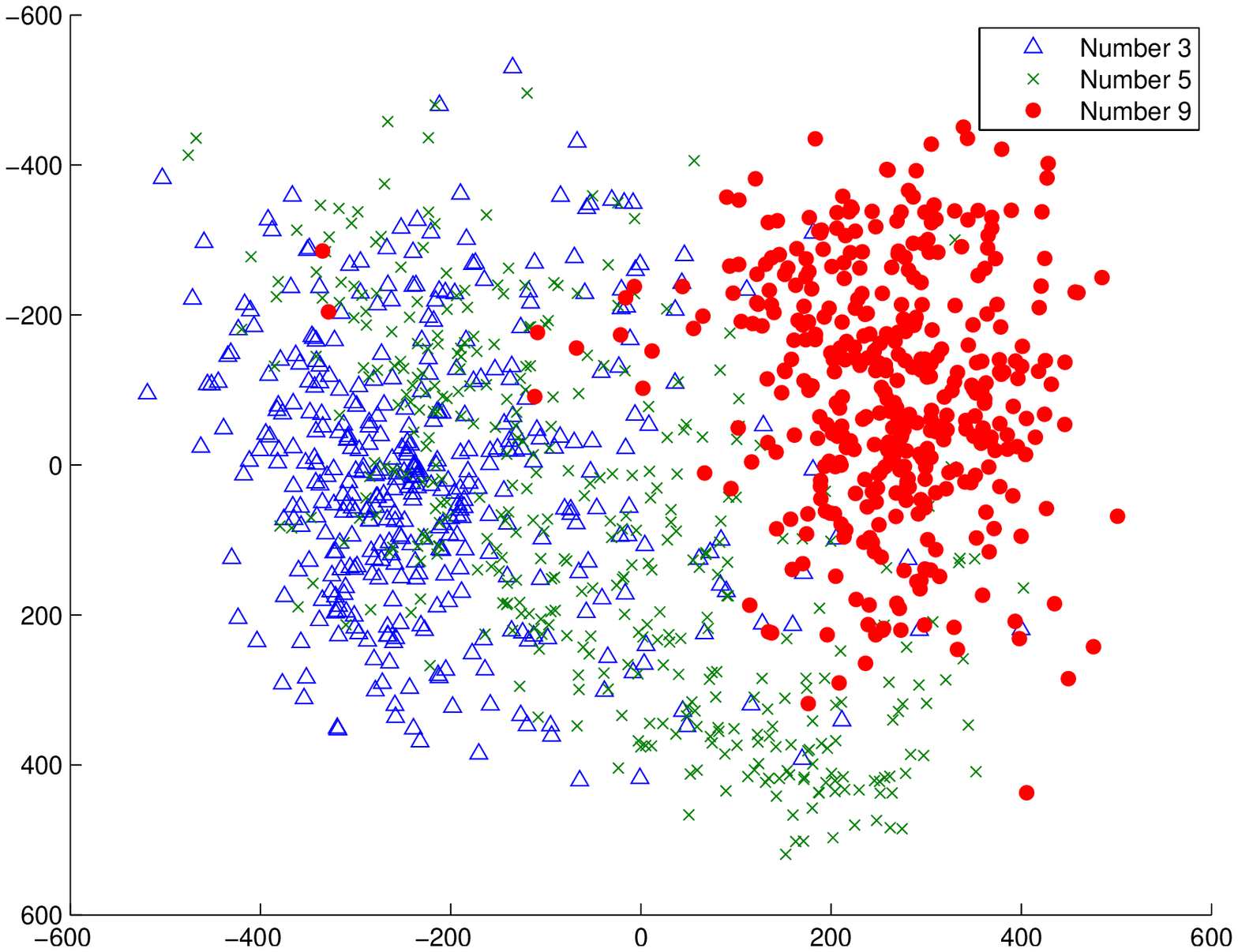}}
  \hspace{0.2in}
  \subfigure[ISOMAP + $\mat{M}^{\textrm{UNI}}$]{
    \label{fig:IsomapGlo} %% label for second subfigure
    \includegraphics[width=2.2in]{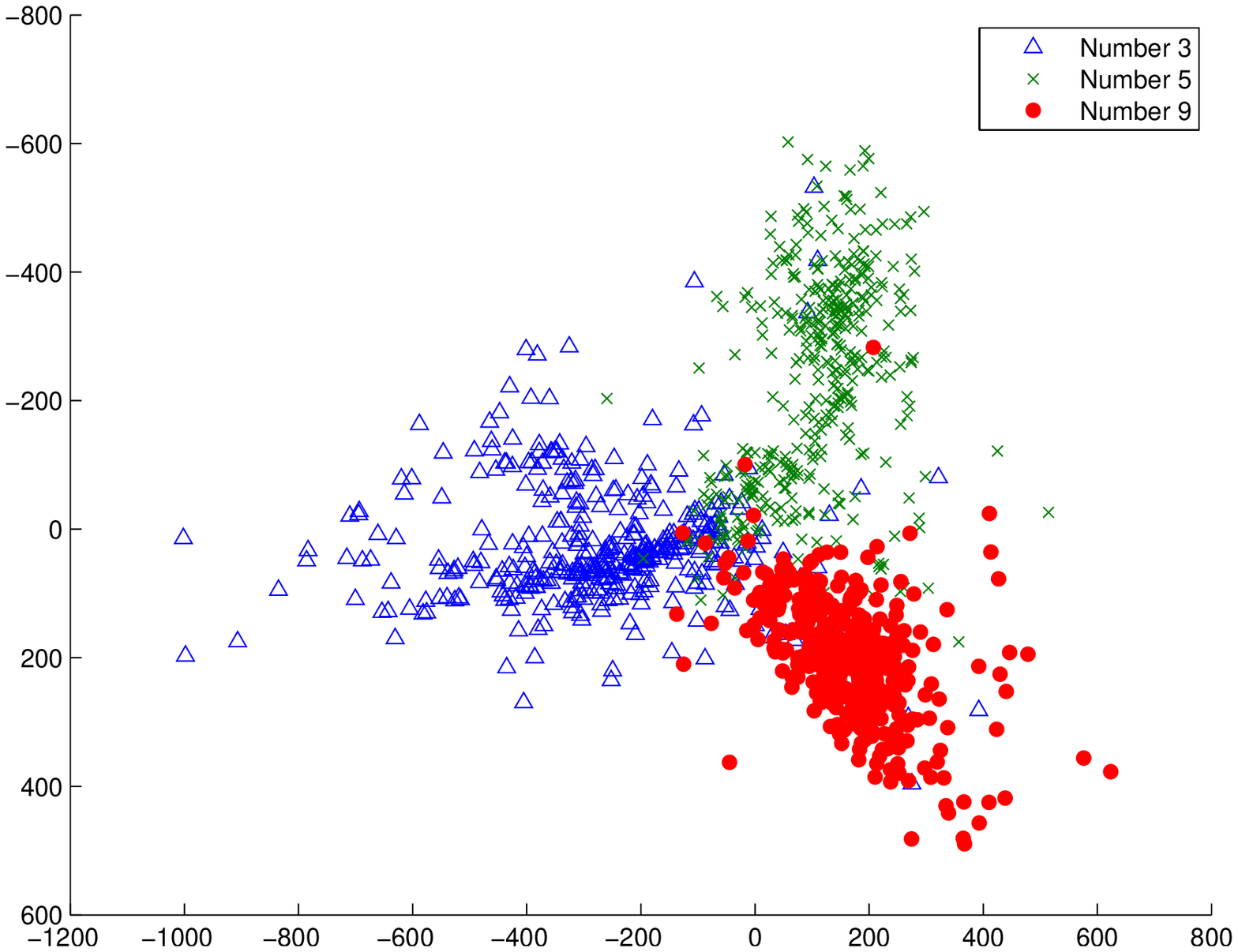}}
  \caption{Isomap embeddings of images of digits, comparing Euclidean
    distance with our method of learning a global metric.}
 \label{fig:Isomap} %% label for entire figure
\end{figure}

In particular, we compare the IsoMap embedding results computed with
the Euclidean metric and the results with the
$\mat{M}^{\textrm{UNI}}$ metric on the MNIST dataset. We selected
400 random samples from different digits `3', `5', and `9', and
resized the images to $7\times 7$. Fig.~\ref{fig:Isomap} plots the
two different low-dimensional embeddings of data samples, colored
according to their digit identities. This clearly shows that
learning a global metric helps to discover a better embedding that
exhibits clear clustering structure among different class
identities.

%% file: discuss.tex
\section{Discussion}
\label{sDisc}

In the context of metric learning, we have proposed several new approaches that can reap the benefits of both discriminative training and generative modeling.  Our method builds upon the connection between a kernel learning framework and using learned positive semidefinite metrics from generative models as base kernels.  Empirical studies validate our algorithms in both improving classification performance across a variety of datasets as well as in computational efficiency in implementation.  Ongoing work includes further investigations into more effective approaches to training nonlinear combinations of learned local metrics in a discriminative manner.

%% file: metric_supplement_content.tex
\section{Proof of Theorem~\ref{thUniform}}
\label{sProof}

\textbf{Theorem~\ref{thUniform}} \textit{Assume the class
conditional distribution $p_c(\vct{x}) = p(\vct{x}|y= c)$ is
Gaussian for every class. Let $\mat{M}_i$ be the local metric
computed with eq.~(\ref{eLocalSDP}), minimizing the bias term in the
space of $\vct{x}$. Then, the uniformly combined metric
$\mat{M}^{\textrm{UNI}}$ of eq.~(\ref{eUniform}) indues a linear
transformation $\vct{z} = \mat{L}\vct{x}$ where $\mat{L}\T\mat{L} =
\mat{M}$. Furthermore, let $\mat{Q}_i$ denote the local metric
computed in the space of $\vct{z}$ under the new class conditional
distribution $p_c(\vct{z})$. We have,
\begin{equation}
\sum_{i=1}^{\cst{N}} \mat{Q}_i \propto \mat{I}
\end{equation}
where $\mat{I}$ is the identity matrix. }

\textit{Proof.} Let $\mat{\Phi}_i$ denote the matrix characterizing the bias term on
$\vct{x}_i$ in the original space (cf. eq.~(\ref{eq:MetricEffect})).  For multiway classification, the
matrix is given by
\begin{equation}
\mat{\Phi} = \sum_{c=1}^{\cst{C}} \nabla\nabla p_c(\vct{x}) \left(
\sum_{c' \ne c} p_{c'}^2(\vct{x})- p_{c}(\vct{x})\sum_{c' \ne
c}p_{c'}(\vct{x})\right)
\end{equation}
where we have dropped the subscript $i$ for clarity. For Gaussian
class conditional distributions $p_c(\vct{x}) = \mathcal{N}(\vct{x}|
\vct{\mu}_c, \mat{\Sigma}_c)$, the Hessian $\nabla\nabla
p_c(\vct{x})$ is given by
\begin{equation}
\nabla\nabla p_c(\vct{x})  =
p_c(\vct{x})\left[\mat{\Sigma}_c^{-1}(\vct{x}-\vct{\mu}_c)(\vct{x}-\vct{\mu}_c)\T\mat{\Sigma}_c^{-1}
- \mat{\Sigma}_c^{-1}\right]
\end{equation}
Under the linear transformation $\mat{L}\vct{x}$, the matrix for the bias term in
the new space is $\mat{\Psi}_i = \mat{L}^{-1} \mat{\Phi}_i
\mat{L}\TN$. We now establish the relationship between $\mat{Q}_i$,
which satisfies
\begin{equation}
\mathsf{Trace}\left[\mat{Q}_i^{-1}\mat{\Psi}_i\right] = 0,\ \
|\mat{Q}_i|=1,\ \mat{Q}_i \succeq 0 \label{eQ}
\end{equation}
and $\mat{M}_i$, the solution in the original space
\begin{equation}
\mathsf{Trace}\left[\mat{M}_i^{-1}\mat{\Phi}_i\right] = 0,\ \
|\mat{M}_i|=1,\ \mat{M}_i \succeq 0 \label{eM}
\end{equation}
Let $\mat{Q}_i = |\mat{L}|^{2/\cst{D}} \mat{L}^{-1} \mat{M}_i
\mat{L}\TN$, where $\cst{D}$ is the input dimensionality. We claim
$\mat{Q}_i$ is the solution to eq.~(\ref{eQ}) as long as $\mat{M}_i$
is the solution to eq.~(\ref{eM}):
\begin{equation}
\mathsf{Trace}\left[\mat{Q}_i^{-1}\mat{\Psi}_i\right] =
\mathsf{Trace}\left[ \mat{L}\T \mat{M}_i^{-1} \mat{L} \mat{L}^{-1}
\mat{\Phi}_i \mat{L}\TN \right] =
\mathsf{Trace}\left[\mat{M}_i^{-1}\mat{\Phi}_i\right] = 0
\end{equation}
and $|\mat{Q}_i| = |\mat{M}_i| = 1$, $\mat{Q}_i \succeq 0 $.

Thus, we have,
\begin{align}
\sum_i \mat{Q}_i = |\mat{L}|^{2/\cst{D}} \sum_i \mat{L}^{-1}
\mat{M}_i \mat{L}\TN & = \cst{N} |\mat{L}|^{2/\cst{D}}  \mat{L}^{-1}
\left( 1/\cst{N} \sum_i \mat{M}_i \right) \mat{L}\TN \\ & = \cst{N}
|\mat{L}|^{2/\cst{D}} \mat{L}^{-1} \mat{M} \mat{L}\TN\label{equ1}
\end{align}

Let the eigen-decomposition of $\mat{M}$ be $\mat{U}
\mat{\Lambda}\mat{U}\T$, where $\mat{\Lambda} =
\mathsf{diag}(\lambda_1,...,\lambda_{\cst{D}})$ contains all
eigenvalues and $\mat{U}$ contains all eigenvectors. We  set
$\mat{L} = \mat{U} \mat{\Delta} \mat{U}\T$, where $\mat{\Delta} =
\mathsf{diag}( \sqrt{\lambda_1},...,\sqrt{\lambda_{\cst{D}}})$.
Then, $\mat{L}$ is an induced transformation from $\mat{M}$ since
$\mat{L}\T\mat{L}=\mat{M}$. Plugging $\mat{L}$ into eq.~(
\ref{equ1}), we obtain:
\begin{align}
\sum_i \mat{Q}_i = \cst{N} |\mat{L}|^{2/\cst{D}}  \mat{L}^{-1}
\mat{M} \mat{L}\TN = \cst{N} |L|^{2/\cst{D}} \mat{I}  \propto
\mat{I}
\end{align}
$\blacksquare$

\section{Linear combination of local
metrics: other forms}\label{sKDE}

The uniform combination eq.~(\ref{eUniform}) and the resulting
metric $\mat{M}^{\textrm{UNI}}$, is a special case of convex
combination of local metrics. Here, we consider another form of
convex combination, where the combination coefficients (or weights)
are proportional to the probabilistic density of each point. The
densities are estimated by density estimators. Typically, a density
estimator also depends on the metric used: a better metric can often
lead to better estimation of densities. Thus we propose the
 iterative procedure listed in the Algorithm~\ref{aDensity} to \emph{jointly} estimate the
densities and combine local metrics.
\begin{algorithm}
\caption{Density-based Convex Combination of Local Metrics}
\label{aDensity}
\begin{algorithmic}[1]
\STATE Compute the local metric $\mat{M}_i$ for each point
$\vct{x}_i$\\
\FOR{$iter = 1 \textit{ to } \textrm{MAXITER}$} \STATE Estimate the
density $p(\vct{x})$ for each point \STATE Compute the global metric
$\vct{M} = \sum_i p(\vct{x}_i) \mat{M}_i$ \STATE Transform each
point by $\vct{x}_i \leftarrow \mat{L}\vct{x}_i$, where
 $\mat{L}\T\mat{L} = \mat{M}$\ENDFOR \STATE Return
$\mat{M}$
\end{algorithmic}
\end{algorithm}

We have experimented with two types of density estimators:
\begin{itemize}
  \item Kernel Density Estimator (KDE). Given the training
  data $\{\vct{x}_1, ..., \vct{x}_n\}$, the density of a testing
  point $\vct{x}_t$ is defined as $p(\vct{x}_t) = 1/h \sum_{i=1}^n \exp( -\|\vct{x}_t - \vct{x}_i\|_2^2 /
  \sigma^2)$, where $h$ is a normalization constant, and $\sigma$ is
  the bandwidth parameter which is tuned on the validation data (to
  maximize the likelihood).
  \item Gaussian Mixture Model (GMM). The model is built by modeling
  each class as a single Gaussian. Note that the densities calculated from GMM are metric invariant if the class assignments are fixed. To make
    the density metric-dependent, we use the following trick: once the global metric
    is learnt, we use it to re-classify all training data (by $k$NN)
   based on the validation data. After that, we build GMM according to the new
    labels, and reiterate the process.
\end{itemize}

\section{Experimental details on metric learning for supervised learning tasks}
\label{sExpDetail}

\subsection{Linear combination of local metrics}

%Yuan: please provide a bit of details about how discriminative
%training of nonlinear combination of metric is done

We compare eight metric learning methods, including two new methods
described in Section \ref{sKDE}:
\begin{itemize}
\item Euclidean.  We use the identity matrix as a metric to compute distances.
  \item LMNN. We learn a  single metric discriminatively using the large margin nearest neighbor method~\cite{weinberger09distance}, as reviewed in section~\ref{sBackground}.
  \item GLM$^{\textrm{INT}}$.  We learn local metrics using generative techniques~\cite{noh10local}, as reviewed in section~\ref{sBackground}.  We use Gaussian distributions as the class conditional distributions for our generative modeling. We follow the procedure defined in~\cite{noh10local} to interpolate the learned local metrics  with the Euclidean metric when we classify new data points. We \emph{do not} report the results of un-interpolated metrics GLMs as our findings are consistent with the authors of ~\cite{noh10local}. The interpolated metrics have much better performance.
  \item $\mat{M}^{\textrm{UNI}}$. This is our approach of  combining the \emph{un-interpolated} GLMs into a single metric with the uniform combination, described in eq.~(\ref{eUniform}) of section~\ref{sConvex}.
  \item LMNN$_E$. With the same metric of LMNN, this method performs energy-based classification~\cite{weinberger09distance}. Loosely speaking, the energy of a point being assigned to a class $c$ is defined as the differences between two quantities: the (sum of) distances of this point to its nearest neighbor in class $c$ and the (sum of) distances of this point to its nearest neighbor in classes other than $c$. A point is assigned to the class which has the lowest energy. According to ~\cite{weinberger09distance}, energy-based classification sometimes improve performance significantly over purely nearest-neighbor based classification.
  \item $\mat{M}^{\textrm{GMM}}$. Learn the global metric as a weighted combination of local
  metrics. The weight is proportional to the density estimated from
  Gaussian Mixture Model. The number of iterations in the Algorithm~\ref{aDensity} is set to 20.
  \item $\mat{M}^{\textrm{KDE}}$. Learn the global metric as a weighted combination of local
  metrics. The weight is proportional to the density estimated from
  (Gaussian) Kernel Density Estimator. The number of iterations in the Algorithm~\ref{aDensity} is set to 20.
  \end{itemize}

For energy-based classification, we need to set a quantity called
``margin''~\cite{weinberger09distance}. We have used the follow
procedure:
\begin{itemize}
  \item transform all samples by the learnt metric.
  \item for each sample, compute the difference from: a) the distance
to its nearest neighbor in the same class; b) the distance to its
nearest neighbor in other classes.
  \item compute the median of these differences and denote its value as
$\gamma_0$. Consider $\beta \gamma_0$ as candidate margins, where
$\beta$ is a scaling factor tuned on validation set.
\end{itemize}

The error rates (mean and standard error) on small-scale datasets
are listed in Table \ref{error-small}, with ranking information
shown in Table \ref{rank-small}. We can see that on most datasets,
$\mat{M}^{\textrm{UNI}}$ and $\mat{M}_E^{\textrm{UNI}}$ outperform
GLM$^{\textrm{INT}}$, LMNN, LMNN$_E$ and Euclidean. Additionally,
$\mat{M}^{\textrm{KDE}}$ performs better than
$\mat{M}^{\textrm{UNI}}$, $\mat{M}_E^{\textrm{UNI}}$ and
$\mat{M}^{\textrm{GMM}}$ in general. However,
$\mat{M}^{\textrm{KDE}}$ is less efficient than
$\mat{M}^{\textrm{UNI}}$ due to its iterative nature. We plan to
explore more efficient and theoretical-sound combination approach in
our future work.

\begin{table}[t]
\caption{Error rates of misclassification (in \%) on small-scale datasets} \label{error-small}
\begin{center}
\begin{tabular}{|l|c|c|c|c|}\hline
\multicolumn{1}{|c|}{\bf METHOD}  & \multicolumn{4}{c|}{\bf DATASET}
\\ \hline& 3-Normal & Wine & Iris & Heart \\ \hline
Euclidean       &8.17 $\pm$ 0.33      & 4.41 $\pm$ 0.61     & 5.11 $\pm$ 0.69     &22.78 $\pm$ 0.88\\
\hline
LMNN            &4.70 $\pm$ 0.30      & 2.61 $\pm$ 0.38     & 4.78 $\pm$ 0.71     &21.91 $\pm$ 0.84\\
\hline
GLM$^{\textrm{INT}}$           &3.83 $\pm$ 0.18      & 3.96 $\pm$ 0.64     & 3.67 $\pm$ 0.54     &21.60 $\pm$ 0.92\\
\hline
$\mat{M}^{\textrm{UNI}}$         &3.44 $\pm$ 0.25      & \textbf{1.80} $\pm$ 0.40     & 3.33 $\pm$ 0.48     &19.51 $\pm$ 0.90\\
\hline
$\mat{M}^{\textrm{GMM}}$      &\textbf{2.96} $\pm$ 0.21      & 2.97 $\pm$ 0.51     & 3.44 $\pm$ 0.54     &19.69 $\pm$ 0.93\\
\hline
$\mat{M}^{\textrm{KDE}}$      &3.09 $\pm$ 0.22      & 2.07 $\pm$ 0.46     & 3.33 $\pm$ 0.48     &\textbf{18.95} $\pm$ 0.90\\
\hline
LMNN$_E$     &3.70 $\pm$ 0.24      & 2.43 $\pm$ 0.44     & 4.67 $\pm$ 0.54     &20.56 $\pm$ 0.80\\
\hline $\mat{M}_E^{\textrm{UNI}}$   &3.10 $\pm$ 0.23      & 2.52
$\pm$
0.52     & \textbf{3.11} $\pm$ 0.57     &19.26 $\pm$ 0.76\\
 \hline
\hline \multicolumn{1}{|c|}{\bf METHOD}  & \multicolumn{4}{c|}{\bf
DATASET}
\\\hline & Vehicle & Ionosphere & Image & German \\ \hline
Euclidean       &31.29 $\pm$ 0.54      & 15.87 $\pm$ 0.68     & 2.68 $\pm$ 0.13     &27.27 $\pm$ 0.60\\
\hline
LMNN            &21.97 $\pm$ 0.47      & 12.39 $\pm$ 0.57     & 2.14 $\pm$ 0.12     &27.20 $\pm$ 0.58\\
\hline
GLM$^{\textrm{INT}}$           &25.32 $\pm$ 0.65      & \textbf{6.24} $\pm$ 0.52     & 2.89 $\pm$ 0.15     &26.17 $\pm$ 0.47\\
\hline
$\mat{M}^{\textrm{UNI}}$         &17.47 $\pm$ 0.30      & 9.67 $\pm$ 0.48     & 2.47 $\pm$ 0.13     &26.12 $\pm$ 0.62\\
\hline
$\mat{M}^{\textrm{GMM}}$     &18.15 $\pm$ 0.49      & 9.77 $\pm$ 0.49     & 2.27 $\pm$ 0.10     &25.87 $\pm$ 0.68\\
\hline
$\mat{M}^{\textrm{KDE}}$       &17.17 $\pm$ 0.33      & 9.01 $\pm$ 0.49     & \textbf{2.12} $\pm$ 0.12     &26.10 $\pm$ 0.66\\
\hline
LMNN$_E$    &20.37 $\pm$ 0.52      & 11.97 $\pm$ 0.66     & 2.67 $\pm$ 0.13     &26.88 $\pm$ 0.55\\
\hline
$\mat{M}_E^{\textrm{UNI}}$   &\textbf{15.81} $\pm$ 0.52      & 10.80 $\pm$ 0.71     & 3.01 $\pm$ 0.16     &\textbf{25.15} $\pm$ 0.51\\
\hline
\end{tabular}
\end{center}
\end{table}

\begin{table}[t]
\caption{Ranking of different methods} \label{rank-small}
\begin{center}
\begin{tabular}{|l|c|c|c|c|c|c|c|c|c|}\hline
\multicolumn{1}{|c|}{\bf METHOD} & \multicolumn{8}{c|}{\bf DATASET}
& \multicolumn{1}{c|}{\bf Avg.}
\\ \cline{2-9}
&3-Norm. & Iris & Wine & Heart & Vehi. & Ionos. & Image & German  & \textbf{Rank}\\
\hline
Euclidean       & 8 & 8 & 8 & 8 & 8 & 8 & 6 & 8  & 7.75\\
\hline
LMNN            & 7 & 5 & 7 & 7 & 6 & 7 & 2 & 7  & 6\\
\hline
GLM$^{\textrm{INT}}$           & 6 & 7 & 5 & 6 & 7 & 1 & 7 & 5  & 5.5\\
\hline
$\mat{M}^{\textrm{UNI}}$         & 4 & 1 & 2 & 3 & 3 & 3 & 4 & 4  & 3\\
\hline
$\mat{M}^{\textrm{GMM}}$      & 1 & 6 & 4 & 4 & 4 & 4 & 3 & 2  & 3.5\\
\hline
$\mat{M}^{\textrm{KDE}}$      & 2 & 2 & 2 & 1 & 2 & 2 & 1 & 3  & 1.87\\
\hline
LMNN$_E$     & 5 & 3 & 6 & 5 & 5 & 6 & 5 & 6  & 5.13\\
\hline
$\mat{M}_E^{\textrm{UNI}}$   & 3 & 4 & 1 & 2 & 1 & 5 & 8 & 1  & 3.13\\
\hline
\end{tabular}
\end{center}
\end{table}

The error rates (mean and standard error) on large-scale datasets
are shown in Table \ref{error-large}.  $\mat{M}^{\textrm{UNI}}$ and $\mat{M}_E^{\textrm{UNI}}$ generally performs well. In particular, with the low dimensionality of 40, $\mat{M}_E^{\textrm{UNI}}$ reaches almost the same accuracy (error rate: $1.40\%$) as discriminatively trained metrics (LMNN) at a higher dimensionality of 164 (error rate: $1.34\%$).

\begin{table}[t]
\caption{Error rates of misclassification (in \%) on large-scale datasets} \label{error-large}
\begin{center}
\begin{tabular}{|l|c|c|c|c|c|} \hline
\multicolumn{1}{|c|}{\bf METHOD}  & \multicolumn{5}{c|}{\bf DATASET}
\\ \cline{2-6}
& MNIST-40 & MNIST-60 & MNIST-164 & Letters-scaled &
Letters-original
\\ \hline
Euclidean       &2.09 $\pm$ 0.02      & 2.02 $\pm$ 0.02     & 2.16 $\pm$ 0.01     &5.05 $\pm$ 0.07   & 5.25 $\pm$ 0.08       \\
\hline
LMNN            &1.99 $\pm$ 0.03      & 1.84 $\pm$ 0.01     & 1.82 $\pm$ 0.03     &3.91 $\pm$ 0.08   & 3.81 $\pm$ 0.13      \\
\hline
GLM$^{\textrm{INT}}$           &3.75 $\pm$ 0.06      & 3.55 $\pm$ 0.05     & 3.48 $\pm$ 0.09     & 5.55 $\pm$ 0.08  & 5.51 $\pm$ 0.08\\
\hline
$\mat{M}^{\textrm{UNI}}$         &1.93 $\pm$ 0.03      & 1.90 $\pm$ 0.03     & 4.30 $\pm$ 0.04     &3.04 $\pm$ 0.08  & 2.96 $\pm$ 0.09\\
\hline
LMNN$_E$     &1.53 $\pm$ 0.01      & \textbf{1.43} $\pm$ 0.01     & \textbf{1.34} $\pm$ 0.01     &2.98 $\pm$ 0.06  & 2.90 $\pm$ 0.06\\
\hline
$\mat{M}_E^{\textrm{UNI}}$   &\textbf{1.40} $\pm$ 0.01      & 1.44 $\pm$ 0.01     & 3.13 $\pm$ 0.03     &\textbf{2.26} $\pm$ 0.04   & \textbf{2.28}$\pm$ 0.06\\
\hline
\end{tabular}
\end{center}
\end{table}

%\subsection{kernel combination}
%We compare 3 types of methods:
%\begin{itemize}
%  \item \textbf{Identity}. The metric $M$ is the identity matrix. It learns the combination of 15 kernels (due to 15 different bandwidth parameters).
%  \item \textbf{Global}. The metric $M$ is from $\mat{M}^{\textrm{UNI}}$form. It learns the combination of 15 kernels.
%  \item \textbf{Local($k$)}. This method first clusters the data into $k$ clusters
%under the metric from $\mat{M}^{\textrm{UNI}}$form, then for each cluster $c \in
%\{1,...,k\}$, it compute the average of local metrics in that
%cluster (denoted by $G_c$). Then, it learns the combination of
%kernels defined on $\{G_c\}$ (15$k$ kernels in total).
%\end{itemize}

\subsubsection{Training speed} \label{sTime}
The training time of LMNN and $\mat{M}^{\textrm{UNI}}$ on
small-scale and large-scale datasets are given in Table
\ref{time-small} and \ref{time-large}, respectively. Note that the
time reported here is the training time \emph{per tuning} (i.e. run
once with fixed parameters), which does not count the time for
parameter-tuning (required in LMNN). Clearly, on most datasets,
$\mat{M}^{\textrm{UNI}}$ achieves one or two-order-magnitude speedup
over LMNN. It is also interesting to point out that the scale of
features may affect the training time of LMNN significantly (LMNN
runs faster on Letters-original than Letters-scaled), as it can
change the number of active constraints.

\begin{table}[t]
\caption{Training time per tuning (seconds) on small-scale datasets}
\label{time-small}
\begin{center}
\begin{tabular}{|c|c|c|c|c|c|c|c|c|}\hline
& 3-Norm. & Wine & Iris & Heart & Vehi. & Ionos. & Image & German\\
\hline LMNN & 195.3 & 1.9 & 3.5 & 3.2 & 126.6 & 12.1 & 160.7 & 8.2
\\ \hline
$\mat{M}^{\textrm{UNI}}$& 0.3 & 0.03 & 0.04 & 0.06 & 0.5 & 0.3 & 0.8
& 0.3
\\\hline
\end{tabular}
\end{center}
\end{table}

\begin{table}[t]
\caption{Training time per tuning (minutes) on large-scale datasets}
\label{time-large}
\begin{center}
\begin{tabular}{|c|c|c|c|c|c|}\hline
&MNIST-40 & MNIST-60 & MNIST-164 & Letters-scaled & Letters-original \\
\hline LMNN & 42 & 55 & 215 & 14 & 3
\\ \hline
$\mat{M}^{\textrm{UNI}}$& 4 & 7 & 47 & 1 & 1 \\
\hline
\end{tabular}
\end{center}
\end{table}

%Yuan: provide explanation on what is global, local (5) etc. Use the
%text in the main text to introduce.

\subsection{Nonlinear combination of local metrics}

We learn the nonlinear combination of metrics in the framework of
convex combination of nonlinear
kernels~\cite{lanckriet04kernel,rakotomamonjy08simple}. Our baseline
systems learn a kernel in the following form
\begin{equation}
K(\vct{x}, \vct{x}') = \sum_l \alpha_l
\exp\{-\tau_l\twonorm{\vct{x}- \vct{x}'}/\sigma_0^2\}
\label{eBaseline}
\end{equation}
where $\alpha_l$ represents coefficients of convex combination,
$\sigma_0^2$ is a normalization factor to fix the scale of the
kernel. The ``scaled'' (inverse) bandwidth $\tau_l$ takes values
from $[2^{-6}, 2^{-5}, ..., 2^7, 2^{8}]$.

For nonlinear combination of metrics, instead of using all local
metrics, we consider $\cst{P}$ ``regional'' metrics for the sake of
computational efficiency (the regional metrics are obtained by
averaging the local metrics in each cluster). We then use those
regional metrics to compose the desired kernel with the formulation
of eq.~(\ref{eNonlinear}). We use the same scaling scheme for
constructing the baseline eq.~(\ref{eBaseline}) and adjust
$\sigma_0^2$ accordingly (with respect to each different regional
metric).

The combination coefficients are learnt in the framework of
SimpleMKL~\cite{rakotomamonjy08simple}, which minimizes the
empirical risk of the support vector machines (SVM). For simplicity,
let us denote the combined kernel as $\sum_i \alpha_i \mat{K}_i$,
where $\{\mat{K}_i\}$ refers to the set of base kernels. SimpleMKL
essentially solves the following optimization problem for binary
classification (can be extended for multi-way classification using
\emph{one-against-all} or \emph{one-against-one} approaches):
\begin{align}
& \min_{\vct{\alpha}} \max_{\vct{\beta}} \quad  \vct{\beta}^{\T} \vct{e} - \dfrac{1}{2} \vct{\beta}^{\T}  \left(\sum_i \alpha_i \mat{K}_i \right) \vct{\beta} \\
& \mathsf{subject\ to} \quad \vct{\beta} \ge 0, \vct{\beta}^{\T}
\vct{y}= 0; \ \ \vct{\alpha} \ge 0, \alpha\T \vct{e} = 0
\end{align}
where $\vct{\beta}$ is the vector containing SVM dual variables, and
$\vct{e}$ refers to all-one vectors.

The SimpleMKL is an iterative numerical optimization procedure to optimize the kernel combining coefficients $\alpha_l$. The amount of time in finding the optimal solution depends on many factors including the number of base kernels. Thus, its computational cost can be substantive. In this aspect, we view $\mat{M}^{\textrm{UNI}}$, ie, simply averaging local metrics, as a strong contender in both improving performance and computational efficiency.

%Linear combination is the simplest way to combine base
%kernels~\cite{lanckriet04kernel}. The desired kernel function
%$K(\cdot,\cdot)$ is parameterized in the base kernel functions,
%\begin{equation}
%K(\vct{x}_m, \vct{x}_n ) = \sum_{b=1}^{\cst{B}} \beta_b
%K_b(\vct{x}_m, \vct{x}_n)
%\end{equation}
%where the combination coefficients are constrained to be nonnegative
%and sum to one, yielding a convex combination. To identify the
%$\{\beta_b\}$, we minimize the objective function value of
%eq.~(\ref{eSVM}), which is the empirical risk of the SVM using the
%kernel function$K(\cdot, \cdot)$. This gives rise to
%\begin{align}
%& \min_{\vct{\beta}} \max_{\vct{\alpha}} \quad  \vct{\alpha}^T \vct{e} - \dfrac{1}{2} \vct{\alpha}^T  \left(\sum_b \beta_b \mat{K}_b \right) \vct{\alpha}^T \\
%& \mathsf{subject\ to} \quad \vct{\alpha} \ge 0, \vct{\alpha}^T
%\vct{y}= 0; \ \ \vct{\beta} \ge 0, \beta\T \vct{e} = 0
%\end{align}
%where we have slightly abused the notation of $\vct{e}$ to refer to
%two all-one vectors with different dimensionality.

The results on misclassification error rates are shown in
Table~\ref{sup-Nonlinear}, where Local ($\cst{P}$) denotes our
nonlinear combination method with $\cst{P}$ regional metrics. Out of
7 datasets we have experimented, nonlinear combination of metrics
often outperform the baseline, with significant improvement on the
datasets of 3-Normal, Iris and Vehicle. We observe that although
nonlinear combing local metrics is promising, choosing the
\emph{right} number of local kernels is important. (It is often
impractical to combine local kernels defined on all data points due
to the heavy computational cost.)

\begin{table}[t]
\caption{Error rates of misclassification (in \%) with nonlinear
combination of metrics} \label{sup-Nonlinear}
\begin{center}
\begin{tabular}{|c|c|c|c|c|}\hline
\multicolumn{1}{|c|}{\bf METHOD}  & \multicolumn{4}{c|}{\bf DATASET}
\\ \cline{2-5}
& 3-Normal & Wine & Iris & Heart  \\
\hline Baseline & 6.08 $\pm$ 0.27 &     \textbf{2.16} $\pm$ 0.40 &   5.56 $\pm$ 0.65     & 18.09 $\pm$ 0.71\\
\hline Local ($\cst{P} = 1$) & 3.36 $\pm$ 0.26 &   2.43 $\pm$ 0.49     &  3.33 $\pm$ 0.60      & \textbf{17.16} $\pm$ 0.57\\
\hline Local ($\cst{P} = 5$) & 2.72 $\pm$ 0.26 &    4.14 $\pm$ 0.56     & \textbf{2.78} $\pm$ 0.36   & 19.26 $\pm$ 0.82\\
\hline Local ($\cst{P} = 10$) & \textbf{2.31} $\pm$ 0.24 &   3.96 $\pm$ 0.70     & 2.89 $\pm$ 0.47   & 18.95 $\pm$ 0.79\\
\hline \hline \multicolumn{1}{|c|}{\bf METHOD}  &
\multicolumn{4}{c|}{\bf DATASET}
\\ \cline{2-5}
& Vehi. & Ionos. & German &  \\
\cline{1-4} Baseline & 26.53 $\pm$ 0.53  & \textbf{6.90} $\pm$ 0.43   & 25.22 $\pm$ 0.38 &\\
\cline{1-4} Local ($\cst{P} = 1$) & 15.56 $\pm$ 0.45    & 9.01 $\pm$ 0.58   & 24.38 $\pm$ 0.44 & \\
\cline{1-4} Local ($\cst{P} = 5$) & \textbf{15.44} $\pm$ 0.42 & 10.47 $\pm$ 0.65  & 24.30 $\pm$ 0.42 &\\
\cline{1-4} Local ($\cst{P} = 10$) & 15.58 $\pm$ 0.47 & 9.81 $\pm$
0.53   &
\textbf{24.25} $\pm$ 0.40 & \\
\hline
\end{tabular}
\end{center}
\end{table}

\section{Application to unsupervised learning}
\label{sUn}

\begin{table}[t]
\caption{RAND score on small-scale datasets} \label{sup-rand-small}
\begin{center}
\begin{tabular}{|c|c|c|c|c|}\hline
\multicolumn{1}{|c|}{\bf METHOD}  & \multicolumn{4}{c|}{\bf DATASET}
\\ \cline{2-5}
& 3-Normal & Iris & Wine & Heart  \\
\hline $k$-means+Euclidean & 0.528 $\pm$ 0.003 & 0.878 $\pm$ 0.009 & \textbf{0.935} $\pm$ 0.008 & \textbf{0.654} $\pm$ 0.008\\
\hline $k$-means+$\mat{M}^{\textrm{UNI}}$ & \textbf{0.545} $\pm$ 0.008 & \textbf{0.948} $\pm$ 0.006 & 0.930 $\pm$ 0.008 & 0.643 $\pm$ 0.007 \\
\hline \hline \multicolumn{1}{|c|}{\bf METHOD}  &
\multicolumn{4}{c|}{\bf DATASET}
\\ \cline{2-5}
& Vehi. & Ionos. & Image & German  \\
\hline $k$-means+Euclidean & 0.651 $\pm$ 0.001 & 0.565 $\pm$ 0.007 & 0.510 $\pm$ 0.002 & 0.500 $\pm$ 0.0005\\
\hline $k$-means+$\mat{M}^{\textrm{UNI}}$ & \textbf{0.657} $\pm$ 0.002 & \textbf{0.569} $\pm$ 0.008 & \textbf{0.567} $\pm$ 0.006 & 0.500 $\pm$ 0.0005 \\
\hline
\end{tabular}
\end{center}
\end{table}

Many unsupervised learning problems, such as clustering and
dimensionality reduction, also depend on using a proper metric to
calculate distances. We investigate how to apply the supervised
metric learning algorithms to such problems.  The crucial step is to
extract labels for the algorithms to compute better metrics.

In unsupervised clustering, we start with $k$-means clustering with
the Euclidean distance metric. We then treat the cluster labels
\emph{as if} they are ground-truth class labels. We apply the
generative metric learning algorithm to such ``labeled'' data,
compute local metrics and then global metric.  We then apply
$k$-means clustering again using the learnt global metric to compute
distances.  We iterate the process for a few times or until the
cluster labels no longer change. Our empirical study shows that this
simple strategy works well. The algorithm returns with clustering of
higher quality, measured in standard measures, than $k$-means. Note
that being able to have a global metric is essential. Without it, it
is difficult to compare distances measured in different (local)
metrics.

We demonstrate the usage of our global metric in the clustering
problem. We use the small-scale datasets in the previous section,
and try $k$-means clustering with the metric from
$\mat{M}^{\textrm{UNI}}$, as well as with the standard Euclidean
metric. We set the $k$ equal to the number of classes, and
iteratively obtain labels and metrics, as described previously.

The clustering results are measured by the RAND score. It is a
similarity measure between two label sets, where the maximum 1
indicates the two labels sets show the exactly same clustering
results. We calculate the RAND score based on the cluster assignment
returned by $k$-means and the true labels of these datasets.

For unsupervised learning, we find that it is useful to regularize
covariance matrices and interpolate local metrics, before computing
the global metric. Indeed, regularization and interpolation can
prevent overfitting, and generally lead to better clustering
results. To tune these parameters, we split each dataset into a
training, validation and testing set by the ratio of $60/20/20$. We
adopt the following procedure for parameter tuning: for each
parameter combination, learn clusters on the training set, and use
the clusters to cluster the validation set. Then, we compute the
RAND score on the validation set to measure the performance of
current parameter combination. Finally, we use the best tuned
parameter combination to learn the clusters on the training set, and
use them to cluster the testing set. We report the RAND score on the
testing set as an indicator of the model performance.

Table \ref{sup-rand-small} gives the RAND scores of different
methods, averaged over 30 splits. We find that $k$-means +
$\mat{M}^{\textrm{UNI}}$ performs better than $k$-means + Euclidean
on 5 (out of 8) datasets, with significant improvement on Iris and
Image datasets. However, metric learning can also have negative
effect, as revealed on Heart and Wine datasets.

%\begin{table}[t]
%\caption{RAND score on small-scale datasets} \label{rand-small}
%\begin{center}
%\begin{tabular}{|l|c|c|c|c|c|c|c|c|}\hline
%\multicolumn{1}{|c|}{\bf METHOD} & \multicolumn{8}{c|}{\bf DATASET}
%\\\hline
%&3Gaussian & Iris & Wine & Heart & Vehicle & Ionosphere & Image &
%German \\  \hline $k$-means + Euclidean & 72.60 & 87.37 & 94.15 &
%68.33 & 65.68 & 58.89 & 51.96 & 54.51
%\\
%\hline $k$-means + $\mat{M}^{\textrm{UNI}}$form & 74.28 & 96.56 & 94.67 & 68.33 &
%66.17 & 59.38 & 60.39 & 54.51\\
%\hline
%\end{tabular}
%\end{center}
%\end{table}

%\begin{figure}[h]
%\begin{center}
%%\framebox[4.0in]{$\;$}
%\subfigure[1]
%\includegraphics{isomap_plot_classes_359.eps}
%\subfigure[2]
%\includegraphics{2_isomap_GGM_plot_classes_359.eps}
%\end{center}
%\caption{Sample figure caption.}
%\end{figure}

% \begin{table}[t]
% \caption{RAND score on small-scale datasets} \label{rand-small}
% \begin{center}
% \begin{tabular}{lcccc}
% \multicolumn{1}{c}{\bf METHOD}  & \multicolumn{4}{c}{\bf DATASET}
% \\ & 3Gaussian & Wine & Iris & Heart \\ \hline
% Euclidean   & 55.98 $\pm$ 5.94  & 93.08 $\pm$ 0.82 & 84.95 $\pm$ 5.10 & 62.76 $\pm$ 6.80 \\
% Global      & 55.98 $\pm$ 5.94  & 92.48 $\pm$ 0.14 & 96.56 $\pm$ 0.00 & 63.02 $\pm$ 6.94 \\
% \hline \hline \multicolumn{1}{c}{\bf METHOD}  &
% \multicolumn{4}{c}{\bf DATASET}
% \\& Vehicle & Ionosphere & Image & German \\ \hline
% Euclidean  & 65.02 $\pm$ 0.73  & 58.78 $\pm$ 0.12 & 51.96 $\pm$
% 0.00 & 50.93 $\pm$
% 1.76 \\
% Global     & 65.20 $\pm$ 0.54  & 58.18 $\pm$ 0.00 & 60.39 $\pm$ 0.00 & 50.93 $\pm$ 1.76 \\
% \end{tabular}
% \end{center}
% \end{table}